\documentclass[11pt,a4paper]{article}
\usepackage[hyperref]{acl2017}
\usepackage{times}
\usepackage{latexsym}

\usepackage{url}

\aclfinalcopy 

\title{Self-Attentional Models Application \\in Task-Oriented Dialogue Generation Systems}

\author{Mansour Saffar Mehrjardi, Amine Trablesi, Osmar R. Za\"{\i}ane\\
  Department of Computing Science, University of Alberta \\
  {\tt \{saffarme,atrabels,zaiane\}@ualberta.ca} \\}

\date{\today}

\begin{document}
\maketitle
\begin{abstract}
  Self-attentional models are a new paradigm for sequence modelling tasks which differ from common sequence modelling methods, such as recurrence-based and convolution-based sequence learning, in the way that their architecture is only based on the attention mechanism. Self-attentional models have been used in the creation of the state-of-the-art models in many NLP tasks such as neural machine translation, but their usage has not been explored for the task of training end-to-end task-oriented dialogue generation systems yet. In this study, we apply these models on the three different datasets for training task-oriented chatbots. 
  Our finding shows that self-attentional models can 
  be exploited to create
  end-to-end task-oriented chatbots 
  which not only achieve
  higher 
  evaluation scores
  compared to recurrence-based models,
  but also do so
  more efficiently.

\end{abstract}



\section{Introduction}
Task-oriented chatbots are a type of dialogue generation system which tries to help the users accomplish specific tasks, such as booking a restaurant table or buying movie tickets, in a continuous and uninterrupted conversational interface and usually in as few steps as possible. The development of such systems falls into the Conversational AI  domain which is the science of developing agents which are able to communicate with humans in a natural way ~\cite{ram2018conversational}. Digital assistants such as Apple's Siri, Google Assistant, Amazon Alexa, and Alibaba's AliMe are examples of successful chatbots developed by giant companies to engage with their customers.

There are mainly two different ways to create a task-oriented chatbot which are either using set of hand-crafted and carefully-designed rules or use corpus-based method in which the chatbot can be trained with a relatively large corpus of conversational data. Given the abundance of dialogue data, the latter method seems to be a better and a more general approach for developing task-oriented chatbots. The corpus-based method also falls into two main chatbot design architectures which are pipelined and end-to-end architectures~\cite{chen2017survey}. End-to-end chatbots are usually neural networks based~\cite{shang2015neural,dodge2015evaluating,wen2016network,eric2017copy} and thus can be adapted to new domains by training on relevant dialogue datasets for that specific domain. Furthermore, all sequence modelling methods can also be used in training end-to-end task-oriented chatbots. A sequence modelling method receives a sequence as input and predicts another sequence as output. For example in the case of machine translation the input could be a sequence of words in a given language and the output would be a sentence in a second language. In a dialogue system, an utterance is the input and the predicted sequence of words would be the corresponding response. 

Self-attentional models are a new paradigm for sequence modelling tasks which differ from common sequence modelling methods, such as recurrence-based and convolution-based sequence learning, in the way that their architecture is only based on the attention mechanism.
The Transformer~\cite{vaswani2017attention} and Universal Transformer~\cite{dehghani2018universal} models are the first models that entirely rely on the self-attention mechanism for both encoder and decoder, and that is why they are also referred to as a self-attentional models. The Transformer models has produced state-of-the-art results in the task neural machine translation~\cite{vaswani2017attention} and this encouraged us to further investigate this model for the task of training task-oriented chatbots. While in the Transformer model there is no recurrence, it turns out that the recurrence used in RNN models is essential for some tasks in NLP including language understanding tasks and thus the Transformer fails to generalize in those tasks~\cite{dehghani2018universal}. We also investigate the usage of the Universal Transformer for this task to see how it compares to the Transformer model.

We focus on self-attentional sequence modelling for this study and intend to provide an answer for one specific question which is:
\begin{itemize}
\item How effective are self-attentional models for training end-to-end task-oriented chatbots?
\end{itemize}
Our contribution in this study is as follows:
\begin{itemize}
\item We train end-to-end task-oriented chatbots using both self-attentional models and common recurrence-based models used in sequence modelling tasks and compare and analyze the results using different evaluation metrics on three different datasets.
\item We provide insight into how effective are self-attentional models for this task and benchmark the time performance of these models against the recurrence-based sequence modelling methods.
\item We try to quantify the effectiveness of self-attention mechanism in self-attentional models and compare its effect to recurrence-based models for the task of training end-to-end task-oriented chatbots.
\end{itemize}

\section{Related Work}
\subsection{Task-Oriented Chatbots Architectures}
End-to-end architectures are among the most used architectures for research in the field of conversational AI. The advantage of using an end-to-end architecture is that one does not need to explicitly train different components for language understanding and dialogue management and then concatenate them together. Network-based end-to-end task-oriented chatbots as in~\cite{wen2016network,bordes2016learning} try to model the learning task as a policy learning method in which the model learns to output a proper response given the current state of the dialogue. As discussed before, all encoder-decoder sequence modelling methods  can be used for training end-to-end chatbots. Eric and Manning~\shortcite{eric2017copy} use the copy mechanism augmentation on simple recurrent neural sequence modelling and achieve good results in training end-to-end task-oriented chatbots~\cite{gu2016incorporating}. 

Another popular method for training chatbots is based on memory networks. Memory networks augment the neural networks with task-specific
memories which the model can learn to read and write. Memory networks have been used in~\cite{bordes2016learning} for training task-oriented agents in which they store dialogue
context in the memory module, and then the model uses it to select a system response (also stored in the memory module) from a set of candidates. A variation of Key-value memory networks~\cite{miller2016key} has been used in~\cite{eric2017key} for the training task-oriented chatbots which stores the knowledge base in the form of triplets (which is (subject,relation,object) such as (yoga,time,3pm)) in the key-value
memory network and then the model tries to select the most relevant entity from the memory and create a relevant response. This approach makes the
interaction with the knowledge base smoother compared to other models.

Another approach for training end-to-end task-oriented dialogue systems
tries to model the task-oriented dialogue generation in a reinforcement learning approach in which the current state of the conversation is passed to some sequence learning network, and this network decides the action which the chatbot should act upon. End-to-end LSTM based model~\cite{williams2016end}, and the Hybrid Code Networks~\cite{williams2017hybrid} can use both supervised and reinforcement learning approaches for training task-oriented chatbots.

\subsection{Sequence Modelling Methods}
Sequence modelling methods usually fall into recurrence-based, convolution-based, and self-attentional-based methods. In recurrence-based sequence modeling, the words are fed into the model in a sequential way, and the model learns the dependencies between the tokens given the context from the past (and the future in case of bidirectional Recurrent Neural Networks (RNNs))~\cite{goodfellow2016deep}. RNNs and their variations such as Long Short-term Memory (LSTM)~\cite{hochreiter1997long}, and Gated Recurrent Units (GRU)~\cite{cho2014properties} are the most widely used recurrence-based models used in sequence modelling tasks. Convolution-based sequence modelling methods rely on Convolutional Neural Networks (CNN)~\cite{lecun1998gradient} which are mostly used for vision tasks but can also be used for handling sequential data. In CNN-based sequence modelling, multiple CNN layers are stacked on top of each other to give the model the ability to learn long-range dependencies. The stacking of layers in CNNs for sequence modeling allows the model to grow its receptive field, or in other words context size, and thus can
model complex dependencies between different sections of the input sequence~\cite{gehring2017convolutional,yu2015multi}. WaveNet~\shortcite{van2016wavenet}, used in audio synthesis, and ByteNet~\shortcite{kalchbrenner2016neural}, used in machine translation tasks, are examples of models trained using convolution-based sequence modelling.

\section{Models}
We compare the most commonly used recurrence-based models for sequence modelling and contrast them with Transformer and Universal Transformer models. The models that we train are:
\subsection{LSTM and Bi-Directional LSTM}
Long Short-term Memory (LSTM) networks are a special kind of RNN networks which can learn long-term dependencies~\cite{hochreiter1997long}. RNN models suffer from the vanishing gradient problem~\cite{bengio1994learning} which makes it hard for RNN models to learn long-term dependencies. The LSTM model  tackles  this  problem  by  defining  a gating  mechanism  which  introduces input,  output and forget gates, and the model has the ability to decide how much of the previous information it needs to keep and how much of the new information it needs to integrate and thus  this mechanism helps the model keep track of long-term dependencies.

Bi-directional LSTMs~\cite{schuster1997bidirectional} are a variation of LSTMs which proved to give better results for some NLP tasks~\cite{graves2005framewise}. The idea behind a Bi-directional LSTM is to give the network (while  training) the ability to not only look  at  past tokens, like LSTM does, but to future tokens, so the model has access to information both form the past and future. In the case of a task-oriented dialogue generation systems, in some cases, the information needed so that the model learns the dependencies between the tokens, comes from the tokens that are ahead of the current index, and if the model is able to take future 
tokens into accounts it can learn more efficiently.
\subsection{Transformer}\label{sec:transformer-model}
As discussed before, Transformer is the first model that entirely relies on the self-attention mechanism for both the encoder and the decoder. The Transformer uses the self-attention mechanism to learn a representation of a sentence by relating different positions of that sentence. Like many of the sequence modelling methods, Transformer follows the encoder-decoder architecture in which the input is given to the encoder and the results of the encoder is passed to the decoder to create the output sequence. The difference between Transformer (which is a self-attentional model) and other sequence models (such as recurrence-based and convolution-based) is that the encoder and decoder architecture is only based on the self-attention mechanism. The Transformer also uses multi-head attention which intends to give the model the ability to look at different representations of the different positions of both the input (encoder self-attention), output (decoder self-attention) and also between input and output (encoder-decoder attention)~\cite{vaswani2017attention}. It has been used in a variety of NLP tasks such as mathematical language understanding [110], language modeling~\cite{dai2018transformer}, machine translation~\cite{vaswani2017attention}, question answering~\cite{devlin2018bert}, and text summarization~\cite{liu2018generating}.

\subsection{Universal Transformer}
The Universal Transformer model is an encoder-decoder-based sequence-to-sequence model which applies recurrence to the representation of each of the positions of the input and output sequences. The main difference between the RNN recurrence and the Universal Transformer recurrence is that the recurrence used in the Universal Transformer is applied on consecutive representation vectors of each token in the sequence (i.e., over depth) whereas in the RNN models this recurrence is applied on positions of the tokens in the sequence. A variation of the Universal Transformer, called Adaptive Universal Transformer, applies the Adaptive Computation Time (ACT)~\cite{graves2013generating} technique on the Universal Transformer model which makes the model train faster since it saves computation time and also in some cases can increase the model accuracy. The ACT allows the Universal Transformer model to use different recurrence time steps for different tokens.


We know, based on reported evidence that transformers are potent in NLP tasks like translation and question answering. Our aim is to assess the applicability and effectiveness of transformers and universal-transformers in the domain of task-oriented conversational agents. In the next section, we report on experiments to investigate the usage of self-attentional models performance against the aforementioned models for the task of training end-to-end task-oriented chatbots.

\section{Experiments}
We run our experiments on Tesla 960M Graphical Processing Unit (GPU). We evaluated the models using the aforementioned metrics and also applied early stopping (with delta set to 0.1 for 600 training steps).

\subsection{Datasets}
We use three different datasets for training the models. We use the Dialogue State Tracking Competition 2 (DSTC2) dataset~\cite{williams2013dialog} which is the most widely used dataset for research on task-oriented chatbots.
We also used two other datasets recently open-sourced by Google Research \cite{shah2018building} which are M2M-sim-M (dataset in movie domain) and M2M-sim-R (dataset in restaurant domain)\footnote{https://github.com/google-research-datasets/simulated-dialogue}. M2M stands for Machines Talking to Machines which refers to the framework with which these two datasets were created. In this framework, dialogues are created via dialogue self-play and later augmented via crowdsourcing. We trained on our models on different datasets in order to make sure the results are not corpus-biased. Table \ref{dataset-info} shows the statistics of these three datasets which we will use to train and evaluate the models.

\begin{table}[ht]
\begin{center}
\begin{tabular}{|c|c|c|c|c|}
\hline \bf Dataset & \bf Num. of Slots & \bf Train & \bf Dev & \bf Test \\\hline
DSTC2 & 8 & 1618 & 1117 & 500\\
\hline
M2M-R & 9 & 1116 & 349 & 775\\
\hline
M2M-M &  5 & 384 & 120 & 264\\
\hline
\end{tabular}
\end{center}
\caption{\label{dataset-info}Statistics of DSTC2, M2M-R, and M2M-M Datasets}
\end{table}

The M2M dataset has more diversity in both language and dialogue flow compared to the the commonly used DSTC2 dataset which makes it appealing for the task of creating task-oriented chatbots. This is also the reason that we decided to use M2M dataset in our experiments to see how well models can handle a more diversed dataset.


\subsubsection{Dataset Preparation}
We followed
the data preparation process used for feeding the conversation history into the encoder-decoder as in~\cite{eric2017copy}. Consider a sample dialogue $D$ in the corpus which consists of a number of turns exchanged between the user and the system. $D$ can be represented as ${(u_1, s_1),(u_2, s_2), ...,(u_k, s_k)}$ where $k$ is the number of turns in this
dialogue. At each time step in the conversation, we encode the conversation turns up to that time step, which is the context of the dialogue so far, and the
system response after that time step will be used as the target. For example, given we are processing the conversation at time step $i$, the context of the
conversation so far would be ${(u_1, s_1, u_2, s_2, ..., u_i)}$ and the model has to learn
to output ${(s_i)}$ as the target.

\begin{table*}
\centering
\begin{tabular}{llcccc}
  Dataset Split & Model & BLEU & Per Turn. Acc & Per Diag. Acc & Entity F1\\
  \hline
test & LSTM (bs=1) & 5.75 & 17.70 & 0.0 & 5.63\\
 &  LSTM + Attention (bs=2) & 30.84 & 18.08 & 0.15 & 32.16\\
 & Bi-LSTM (bs=2) & 30.38 & 18.04 & 0.0 & 24.34\\
 & Bi-LSTM + Attention (bs=2) & 38.64 & 26.04 & 0.62 & 43.52\\
 & Transformer (bs=2) & \bf{51.83} & \bf{39.02} & \bf{1.7} & \bf{64.20}\\
 & UT (bs=2) & 44.93 & 36.62 & 1.08 & 57.98\\
 & UT + ACT (bs=2) & 39.40 & 30.00 & 0.15 & 61.49\\
 \hline
development & LSTM & 16.13 & 10.33 & 0.0 & 6.54\\
 &  LSTM + Attention & 31.05 & 18.68 & 0.31 & 32.59\\
 & Bi-LSTM & 30.92 & 19.07 & 0.31 & 25.91\\
 & Bi-LSTM + Attention & 39.12 & 27.28 & \bf{0.96} & 44.15\\
 & Transformer & \bf{54.18} & \bf{41.09} & 0.62 & \bf{66.02}\\
 & UT & 47.95 & 39.01 & 0.31 & 61.27\\
 & UT + ACT & 39.27 & 29.30 & 0.31 & 62.50\\
\end{tabular}
\caption{{\label{comparison-table-dstc}}Evaluation of Models on DSTC2 dataset for both test and development datasets (bs: shows the best beam size in inference; UT: Universal Transformers)}
\end{table*}

\subsection{Training}
We used the tensor2tensor library~\cite{vaswani2018tensor2tensor} in our experiments for training and evaluation of sequence modeling methods. We use Adam optimizer~\cite{kingma2014adam} for training the models. We set $\beta_1=0.9$, $\beta_2=0.997$, and $\epsilon=1e-9$ for the Adam optimizer and started with learning rate of 0.2 with noam learning rate decay schema~\cite{vaswani2017attention}. In order to avoid overfitting, we use dropout~\cite{srivastava2014dropout} with dropout chosen from [0.7-0.9] range. We also conducted early stopping~\cite{goodfellow2016deep} to avoid overfitting in our experiments as the regularization methods. We set the batch size to 4096, hidden size to 128, and the embedding size to 128 for all the models. We also used grid search for hyperparameter tuning for all of the trained models. Details of our training and hyperparameter tuning and the code for reproducing the results can be found in the \textit{chatbot-exp github repository}\footnote{https://github.com/msaffarm/chatbot-exp}.

\subsection{Inference}
In the inference time, there are mainly two methods for decoding which are greedy and beam search~\cite{freitag2017beam}.
Beam search has been proved to be an essential part in generative NLP task such as neural machine translation~\cite{wu2016google}. In the case of dialogue generation systems, beam search could help alleviate the problem of having many possible valid outputs which do not match with the target but are valid and sensible outputs. Consider the case in which a task-oriented chatbot, trained for a restaurant reservation task, in response to the user utterance  \textit{``Persian food''}, generates the response \textit{``what time and day would you like the reservation for?''} but the target defined for the system is \textit{``would you like a fancy restaurant?''}. The response generated by the chatbot is a valid response which asks the user about other possible entities but does not match with the defined target.

We try to alleviate this problem in inference time by applying the beam search technique with a different beam size
$\alpha \in \{1, 2, 4\}$ and pick the best result based on the BLEU score. Note that when $\alpha = 1$, we are using the original greedy search method for the generation task.

\begin{table*}
\centering
\begin{tabular}{llcccc}
  Dataset Split & Model & BLEU & Per Turn. Acc & Per Diag. Acc & Entity F1\\
  \hline
M2M-R & LSTM(bs=2) & 6.00 & \bf{2.3} & 0.0 & 7.99\\
(test)&  LSTM+Att.(bs=1) & 7.9 & 1.84 & 0.0 & 16.77\\
 & Bi-LSTM(bs=1) & 8.15 & 1.8 & 0.0 & 19.61\\
 & Bi-LSTM+Att.(bs=1) & 8.3 & 0.97 & 0.0 & 24.12\\
 & Transformer(bs=1) & \bf{10.28} & 1.76 & 0.0 & \bf{36.92}\\
 & UT(bs=2) & 9.15 & 1.88 & 0.0 & 25.44\\
 & UT+ACT(bs=2) & 8.54 & 1.43 & 0.0 & 23.12\\
 \hline
M2M-M & LSTM(bs=4) & 7.7 & \bf{3.36} & 0.0 & 31.07\\
(test) &  LSTM+Att.(bs=2) & 8.3 & 3.27 & 0.0 & 31.18\\
 & Bi-LSTM(bs=2) & 9.6 & 2.09 & 0.0 & 28.09\\
 & Bi-LSTM+Att.(bs=2) & 10.62 & 2.54 & 0.0 & 32.43\\
 & Transformer(bs=1) & \bf{11.95} & 2.36 & 0.0 & \bf{39.89}\\
 & UT(bs=2) & 10.87 & 3.15 & 0.0 & 34.15\\
 & UT+ACT(bs=2) & 10.48 & 2.46 & 0.0 & 32.76\\
\end{tabular}
\caption{{\label{comparison-table-m2m}}Evaluation of models on M2M restaurant (M2M-R) and movie (M2M-M) dataset for test datasets (bs: The best beam size in inference; UT: Universal Transformers)}
\end{table*}

\subsection{Evaluation Measures}
{\bf BLEU}: We use the Bilingual Evaluation Understudy (BLEU)~\cite{papineni2002bleu} metric which is commonly used in machine translation tasks. The BLEU metric can be used to evaluate dialogue generation models as in~\cite{eric2017copy,li2015diversity}. The BLEU metric is a word-overlap metric which computes the co-occurrence of N-grams in the reference and the generated response and also applies the brevity penalty which tries to penalize far too short responses which are usually not desired in task-oriented chatbots. We compute the BLEU score using all generated responses of our systems.

{\bf Per-turn Accuracy}: Per-turn accuracy measures the similarity of the system generated response versus
the target response. Eric and Manning~\shortcite{eric2017copy} used this metric to evaluate their systems in which they considered their response to be correct if all tokens in the system generated response matched the corresponding token in the target response. This metric is a little bit harsh, and the results may be low since all the tokens in the generated response have to be exactly in the same position as in the target response.

{\bf Per-Dialogue Accuracy}: We calculate per-dialogue accuracy as used in~\cite{bordes2016learning,eric2017copy}. For this metric, we consider all the system generated responses and compare them to the target responses.
A dialogue is considered to be true if all the turns in the system generated
responses 
match the corresponding turns in the target responses. Note that this is a very strict metric in which all the utterances in the dialogue should be the same as the target and in the right order. 

{\bf F1-Entity Score}: Datasets used in task-oriented chores have a set of entities which represent user preferences. For example, in the restaurant domain chatbots common entities are
meal, restaurant name, date, time and the number of people (these are usually the required entities which are crucial for making reservations, but there could be optional entities such as location or rating). Each target response has
a set of entities which the system asks or informs the user about. Our models have to be able to discern these specific entities and inject them into the generated response. To evaluate our models we could use named-entity recognition evaluation metrics~\cite{jiang2016evaluating}. The F1 score is the most commonly used metric used for the evaluation of named-entity recognition models which is the harmonic
average of precision and recall of the model. We calculate this metric by micro-averaging over all the system generated responses.

\section{Results and Discussion}
\subsection{Comparison of Models}
The results of running the experiments for the aforementioned models is shown in Table~\ref{comparison-table-dstc} for the DSTC2 dataset and in Table~\ref{comparison-table-m2m} for the M2M datasets. The bold numbers show the best performing model in each of the evaluation metrics. As discussed before, for each model we use different beam sizes (bs) in inference time and report the best one. Our findings in Table~\ref{comparison-table-dstc} show that self-attentional models outperform common recurrence-based sequence modelling methods in the BLEU, Per-turn accuracy, and entity F1 score.
The reduction in the evalution numbers for the M2M dataset and in our investigation of the trained model we found that this considerable reduction is due to the fact that the diversity of M2M dataset is considerably more compared to DSTC2 dataset while the traning corpus size is smaller.

\subsection{Time Performance Comparison}
Table~\ref{time-conv-table} shows the time performance of the models trained on DSTC2 dataset. Note that in order to get a fair time performance comparison, we trained the models with the same batch size (4096) and on the same GPU. These numbers are for the best performing model (in terms of evaluation loss and selected using the early stopping method) for each of the sequence modelling methods. Time to Convergence (T2C) shows the approximate time that the model was trained to converge. We also show the loss in the development set for that specific checkpoint.

\begin{table}[ht]
\begin{center}
\begin{tabular}{|c|c|c|}
\hline \bf Model & \bf T2C (sec) & \bf Dev Loss\\\hline
LSTM & 1100 & 0.89 \\
\hline
LSTM+Att & 1305 & 0.62 \\
\hline
Bi-LSTM & 1865 & 0.60 \\
\hline
Bi-LSTM+Att & 2120 & 0.49 \\
\hline
Transformer & \bf{612} & \bf{0.31} \\
\hline
UT & 1939 & 0.36 \\
\hline
UT+ACT & 665 & 0.33 \\
\hline
\end{tabular}
\end{center}
\caption{\label{time-conv-table}Comparison of convergence performance of the models}
\end{table}

\subsection{Effect of (Self-)Attention Mechanism}\label{sec:effect-of-self-attention}
As discussed before in Section~\ref{sec:transformer-model}, self-attentional models rely on the self-attention mechanism for sequence modelling. Recurrence-based models such as LSTM and Bi-LSTM can also be augmented in order to increase their performance, as evident in Table~\ref{comparison-table-dstc} which shows the increase in the performance of both LSTM and Bi-LSTM when augmented with an attention mechanism. This leads to the question whether we can increase the performance of recurrence-based models by adding multiple attention heads, similar to the multi-head self-attention mechanism used in self-attentional models, and outperform the self-attentional models. 

To investigate this question, we ran a number of experiments in which we added multiple attention heads on top of Bi-LSTM model and also tried a different number of self-attention heads in self-attentional models in order to compare their performance for this specific task. Table~\ref{self-attention-effect-table} shows the results of these experiments. Note that the models in Table~\ref{self-attention-effect-table} are actually the best models that we found in our experiments on DSTC2 dataset and we only changed one parameter for each of them, i.e. the number of attention heads in the recurrence-based models and the number of self-attention heads in the self-attentional models, keeping all other parameters unchanged. We also report the results of models with beam size of 2 in inference time. We increased the number of attention heads in the Bi-LSTM model up to 64 heads to see its performance change. Note that increasing the number of attention heads makes the training time intractable and time consuming while the model size would increase significantly as shown in Table~\ref{time-conv-table-second}. Furthermore, by observing the results of the Bi-LSTM+Att model in Table~\ref{self-attention-effect-table} (both test and development set) we can see that Bi-LSTM performance decreases and thus there is no need to increase the attention heads further.

Our findings in Table~\ref{self-attention-effect-table} show that the self-attention mechanism can outperform recurrence-based models even if the recurrence-based models have multiple attention heads. The Bi-LSTM model with 64 attention heads cannot beat the best Trasnformer model with NH=4 and also its results are very close to the Transformer model with NH=1. This observation clearly depicts the power of self-attentional based models and demonstrates that the attention mechanism used in self-attentional models as the backbone for learning, outperforms recurrence-based models even if they are augmented with multiple attention heads.

\begin{table}[ht]
\begin{center}
\begin{tabular}{|c|c|c|}
\hline \bf Model & \bf T2C (sec) & \bf Dev Loss\\\hline
Bi-LSTM+Att.[NH=1] & 2120 & 0.49 \\
\hline
Bi-LSTM+Att.[NH=4] & 3098 & 0.47 \\
\hline
Bi-LSTM+Att.[NH=8] & 3530 & 0.44\\
\hline
Bi-LSTM+Att.[NH=16] & 3856 & 0.44 \\
\hline
Bi-LSTM+Att.[NH=32] & 7320 & 0.36 \\
\hline
Bi-LSTM+Att.[NH=64] & \bf{9874} & \bf{0.38} \\
\hline
Transformer[NH=1] & \bf{375} & \bf{0.33} \\
\hline
Transformer[NH=4] & \bf{612} & \bf{0.31} \\
\hline
Transformer[NH=8] & 476 & 0.31 \\
\hline
\end{tabular}
\end{center}
\caption{\label{time-conv-table-second}Comparison of convergence performance of the models}
\end{table}

\begin{table*}
\centering
\begin{tabular}{llcccc}
 Dataset Split & Model & BLEU & Per-Turn Acc& Per-Diag Acc& Entity F1\\
\hline
 test & Bi-LSTM+Att.[NH=1] & 38.64 & 26.04 & 0.62 & 43.52\\
 & Bi-LSTM+Att.[NH=4] & 42.23 & 29.01 & 0.92 & 48.06\\
 & Bi-LSTM+Att.[NH=8] & 42.61 & 28.18 & 0.77 & 49.90\\
 & \bf{Bi-LSTM+Att.[NH=16]} & \bf{43.11} & \bf{30.34} & \bf{0.61} & \bf{50.87}\\ 
  & \bf{Bi-LSTM+Att.[NH=32]} & \bf{48.62} & \bf{36.46} & \bf{1.85} & \bf{59.8}\\ 
 & \bf{Bi-LSTM+Att.[NH=64]} & \bf{47.33} & \bf{33.17} & \bf{1.23} & \bf{56.49}\\ 
 & \bf{Transformer[NH=1]} & \bf{45.90} & \bf{36.64} & \bf{1.7} & \bf{57.55}\\
& \bf{Transformer[NH=4]} & \bf{51.83} & \bf{39.02} & \bf{1.7} & \bf{64.20}\\
 & Transformer[NH=8] & 51.37 & 39.45 & 3.24 & 62.38\\
 & UT[NH=1] & 43.02 & 31.20 & 1.54 & 60.10\\
 & UT[NH=8] & 48.17 & 35.76 & 2.93 & 61.56\\
 & UT+ACT[NH=1] & 34.98 & 25.66 & 0.46 & 51.32\\
 & UT+ACT[NH=8] & 36.29 & 24.97 & 0.31 & 55.27\\
 \hline
 development & Bi-LSTM+Att.[NH=1] & 39.12 & 27.28 & 0.96 & 44.15\\
 & Bi-LSTM+Att.[NH=4] & 40.47 & 27.64 & 0.93 & 48.10\\
 & Bi-LSTM+Att. [NH=8] & 42.78 & 28.36 & 0.31 & 50.05\\ 
 & \bf{Bi-LSTM+Att.[NH=16]} & \bf{42.88} & \bf{30.36} & \bf{0.93} & \bf{52.09}\\ 
 & \bf{Bi-LSTM+Att.[NH=32]} & \bf{49.36} & \bf{38.24} & \bf{0.61} & \bf{61.26}\\ 
 & \bf{Bi-LSTM+Att.[NH=64]} & \bf{47.28} & \bf{33.12} & \bf{0.93} & \bf{56.86}\\ 
 & \bf{Transformer[NH=1]} & \bf{47.86} & \bf{38.33} & \bf{1.85} & \bf{60.37}\\
 & \bf{Transformer[NH=4]} & \bf{54.18} & \bf{41.09} & \bf{0.62} & \bf{66.02}\\
 & Transformer[NH=8] & 51.54 & 39.42 & 1.54 & 63.56\\
 & UT[NH=1] & 43.01 & 32.12 & 1.58 & 60.42\\
 & UT[NH=8] & 47.89 & 35.57 & 1.23 & 61.33\\
 & UT+ACT[NH=1] & 35.74 & 26.46 & 0.31 & 52.71\\
 & UT+ACT[NH=8] & 38.95 & 27.10 & 0.31 & 57.02\\
\hline\end{tabular}
\caption{{\label{self-attention-effect-table}} Evaluation of effect of self-attention mechanism using DSTC2 dataset (Att: Attetnion mechanism; UT: Universal Transformers; ACT: Adaptive Computation Time; NH: Number of attention heads)}
\end{table*}

\section{Conclusion and Future Work}
We have determined that Transformers and Universal-Transformers are indeed effective at generating appropriate responses in task-oriented chatbot systems. In actuality, their performance is even better than the typically used deep learning architectures.
Our findings in Table~\ref{comparison-table-dstc} show that self-attentional models outperform common recurrence-based sequence modelling methods in the BLEU, Per-turn accuracy, and entity F1 score. The results of the Transformer model beats all other models in all of the evaluation metrics. Also, comparing the result of LSTM and LSTM with attention mechanism as well as the Bi-LSTM with Bi-LSTM with attention mechanism, it can be observed in the results that adding the attention mechanism can increase the performance of the models. Comparing the results of self-attentional models shows that the Transformer model outperforms the other self-attentional models, while the Universal Transformer model gives reasonably good results.

In future work, it would be interesting to compare the performance of self-attentional models (specifically the winning Transformer model) against other end-to-end architectures such as the Memory Augmented Networks.

\bibliography{acl2017}
\bibliographystyle{acl_natbib}

\appendix






\end{document}